\documentclass[twoside,11pt,dvipsnames]{article}
\usepackage[preprint]{tmlr}

\usepackage{amsmath,amsfonts,bm}

\def\eqref#1{equation~\ref{#1}}

\def\1{\bm{1}}

\DeclareMathAlphabet{\mathsfit}{\encodingdefault}{\sfdefault}{m}{sl}
\SetMathAlphabet{\mathsfit}{bold}{\encodingdefault}{\sfdefault}{bx}{n}

\usepackage{threeparttable}    
\usepackage{multirow}   
\usepackage{tabularx}
\usepackage{booktabs}

\usepackage{tikz}
\usetikzlibrary{arrows.meta, positioning, calc}

\usepackage{microtype}
\usepackage{subfigure}
\usepackage{adjustbox}

\usepackage{amsmath}
\usepackage{amssymb}
\usepackage{mathtools}
\usepackage{xspace}
\usepackage{url}
\usepackage{graphicx}
\usepackage{algorithmic}
\usetikzlibrary{tikzmark,calc,positioning,arrows.meta,decorations,decorations.pathmorphing,decorations.pathreplacing}
\usepackage{colortbl}
\usepackage{pifont} 
\usepackage{makecell}
\usepackage{fancyhdr}
\usepackage[normalem]{ulem}
\usepackage{enumitem}

\usepackage{listings}
\usepackage[most]{tcolorbox}
\usepackage{xcolor}

\definecolor{codebg}{RGB}{245,245,244}
\definecolor{coderule}{RGB}{220,220,220}
\definecolor{codekw}{RGB}{60,120,180}
\definecolor{codecom}{RGB}{90,90,90}
\definecolor{codestr}{RGB}{180,80,120}
\definecolor{codenum}{RGB}{180,180,180}
\usepackage{needspace}

\lstdefinestyle{papercode}{
  language=Python,
  numbers=left,
  numberstyle=\tiny\color{codenum},
  stepnumber=1,
  numbersep=0.8em,
  backgroundcolor=\color{codebg},
  frame=shadowbox,
  rulesepcolor=\color{coderule},
  frameround=tttt,
  basicstyle=\ttfamily\small,
  keywordstyle=\color{codekw},
  commentstyle=\color{codecom},
  stringstyle=\color{codestr},
  showstringspaces=false,
  breaklines=true,
  breakatwhitespace=false,
  columns=fullflexible,
  keepspaces=true,
  xleftmargin=0.8em,
  xrightmargin=0.2em,
  aboveskip=0.6em,
  belowskip=0.6em
}

\NewDocumentCommand{\jintao}{ mO{} }{\textcolor{blue}{\textsuperscript{\textit{JT}}\textsf{\small[#1]}}}
\NewDocumentCommand{\xxm}{ mO{} }{\textcolor{blue}{\textsuperscript{\textit{XXM}}\textsf{\small[#1]}}}\NewDocumentCommand{\chunyu}{ mO{} }{\textcolor{red}{\textsuperscript{\textit{CY}}\textsf{\small[#1]}}}
\NewDocumentCommand{\rundong}{ mO{} }{\textcolor{Green}{\textsuperscript{\textit{RD}}\textsf{\small[#1]}}}
\NewDocumentCommand{\ziteng}{ mO{} }{\textcolor{purple}{\textsuperscript{\textit{ZT}}\textsf{\small[#1]}}}
\NewDocumentCommand{\hyz}{ mO{} }{\textcolor{orange}Yellow{\textsuperscript{\textit{hyz}}\textsf{\small[#1]}}}
\NewDocumentCommand{\plz}{ mO{} }{\textcolor{cyan}{\textsuperscript{\textit{PL}}\textsf{\small[#1]}}}
\NewDocumentCommand{\jia}{ mO{} }{\textcolor{magenta}{\textsuperscript{\textit{JIA}}\textsf{\small[#1]}}}
\NewDocumentCommand{\tianyu}{ mO{} }{\textcolor{blue}{\textsuperscript{\textit{TY}}\textsf{\small[#1]}}}
\NewDocumentCommand{\xingyang}{ mO{} }{\textcolor{teal}{\textsuperscript{\textit{XY}}\textsf{\small[#1]}}}

\definecolor{deepgreen}{rgb}{0.0, 0.5, 0.0}  
\definecolor{deepred}{rgb}{0.6, 0.0, 0.0}

\usepackage{hyperref}
\title{
    KernelBenchX: A Comprehensive Benchmark for Evaluating LLM-Generated GPU Kernels
}

\author{\small
Han Wang\thanks{Equal contribution.} \quad
Jintao Zhang\footnotemark[1] \quad
Kai Jiang \quad
Haoxu Wang \quad
Jianfei Chen \quad
Jun Zhu
}

\newcommand{\oursys}{\texttt{KernelBenchX}\xspace}

\usepackage{amsthm}

\begin{document}

\maketitle
\begin{center}
\texttt{\href{https://github.com/BonnieW05/KernelBenchX}{https://github.com/BonnieW05/KernelBenchX}}
\end{center}

\begin{abstract}
LLM-based Triton kernel generation has attracted significant interest, yet a fundamental empirical question remains unanswered: where does this capability break down, and why? We present \oursys, a benchmark designed to answer this question through category-aware evaluation of correctness and hardware efficiency across 176 tasks in 15 categories. 
Our systematic comparison of five representative methods yields three main findings. 
First, task structure determines correctness more than method design. Category explains nearly three times more variance in semantic correctness than method (9.4\% vs.\ 3.3\% explained deviance), and 72\% of Fusion tasks fail across all five methods while Math tasks are solved consistently. 
Second, iterative refinement improves correctness, but not performance. Across GEAK iterations, compile rate rises from 52.3\% to 68.8\% while average speedup declines from $1.58\times$ to $1.44\times$; newly rescued kernels consistently underperform persistently correct ones ($1.16\times$ vs.\ $1.58\times$ speedup in round~0$\to$1). 
Third, correctness does not imply efficiency. 46.6\% of correct kernels are slower than the PyTorch eager baseline, and cross-hardware speedup variance reaches $21.4\times$. 
Besides, quantization remains completely unsolved (0/30 successes) despite non-trivial compilation rates, revealing systematic misunderstanding of numerical computation contracts rather than surface-level syntax errors. 
These findings suggest that future progress depends on handling global coordination, explicitly modeling numerical precision, and incorporating hardware efficiency into generation. The code is available at \url{https://github.com/BonnieW05/KernelBenchX}
\end{abstract}


\section{Introduction}

\paragraph{Background.}
GPU kernel efficiency has become a central bottleneck in large-scale machine learning workloads~\cite{shah2024flashattention, zhang2025turbodiffusion, zhangefficient, zhangsageattention}. Prior work such as DeepSeek-V3~\cite{liu2024deepseek} demonstrates that at scale, competitive performance depends not only on model architecture but critically on kernel efficiency.
This has motivated a growing body of work using LLMs to automate Triton~\cite{tillet2019triton} kernel generation~\cite{yu2026towards}, ranging from training-based methods (AutoTriton~\cite{li2025autotriton}, TritonRL~\cite{woo2025tritonrl}) to agent-based iterative systems (GEAK~\cite{wang2025geak}, STARK~\cite{dong2025stark}) to search-and-reasoning approaches (KernelEvolve~\cite{liao2025kernelevolve}, ReGraph~\cite{gong2025large}).
\nocite{zhang2025spargeattn, zhang2025sla, zhang2026sla2, zhang2024sageattention2, zhang2025sageattention2++, zhang2025sageattention3, zhang2026sagebwd}

Accompanying this methodological progress, several benchmarks have been proposed. KernelBench~\cite{ouyang2025kernelbench} provides multi-level evaluation from operators to end-to-end models. TritonBench~\cite{li2025tritonbench} focuses on Triton kernels with a dual-channel testing framework. MultiKernelBench~\cite{wen2025multikernelbench} adds cross-platform evaluation, and Robust-KBench~\cite{lange2025towards} emphasizes robustness against misleading performance gains.

\paragraph{Limitations.}
Despite recent progress, two fundamental questions about LLM-based kernel generation remain unresolved.
First, the capability boundary is not well described: we do not yet know which types of tasks current methods handle reliably, which consistently fail, and why.
Second, the role of iterative refinement is not well understood: it is unclear whether different strategies improve compilation, correctness, or performance, and to what extent.
However, existing benchmarks are not designed to answer these questions, due to their unresolved task categories, insufficient correctness verification, and limited evaluation of efficiency.

\paragraph{Our Method.}
To address these limitations, we propose \textbf{\oursys}, built on TritonBench-T and extended in three directions: (1) a robust two-stage correctness protocol that rejects implementations passing output comparison by chance; (2) a unified 15-category taxonomy together with quantization and multi-precision task extensions, enabling fine-grained structural analysis; and (3) hardware-efficiency metrics beyond runtime.

\paragraph{Experimental Overview.}

Using \oursys, we conduct a systematic comparison of representative Triton kernel generation methods. Beyond aggregate benchmark results, we analyze method behavior across task categories, correctness outcomes, and efficiency metrics under a unified evaluation pipeline. We also collect error-correction and optimization pairs from the generation process for future training and inference-time improvement.

\paragraph{Contributions.}
Our main contributions are as follows:
\begin{itemize}
    \item We introduce \textbf{\oursys}, a benchmark for Triton kernel generation with category-aware evaluation of correctness and hardware efficiency.
    \item We conduct a systematic comparison of five representative methods under a unified pipeline.
    \item We identify three empirical findings that characterize the capability boundary of LLM-based kernel generation and provide mechanistic analysis for each.
    \item We release error--correction and optimization pairs collected during evaluation to support future training and inference.
\end{itemize}
\section{Preliminary}  
\label{Sec:Preliminary}

\subsection{Hardware Efficiency Metrics}

Hardware efficiency describes how effectively a kernel uses the available hardware resources. Beyond runtime, it provides a more direct view of whether performance is close to the hardware limit.
For a kernel $k$, let $T(k)$ denote the measured runtime, $B(k)$ the total bytes moved, and $F(k)$ the total floating-point operations. We define the achieved bandwidth and achieved throughput as
\begin{equation}
\mathrm{BW}(k)=\frac{B(k)}{T(k)}, \qquad
\mathrm{TP}(k)=\frac{F(k)}{T(k)},
\end{equation}
reported in GB/s and TFLOPS respectively.
To enable comparison across hardware, we normalize these quantities by their corresponding peak values:
\begin{equation}
\mathrm{IOU}(k)=\frac{\mathrm{BW}(k)}{\mathrm{BW}_{\max}}, \qquad
\mathrm{MFU}(k)=\frac{\mathrm{TP}(k)}{\mathrm{TP}_{\max}}.
\end{equation}
IOU measures memory bandwidth utilization, while MFU measures compute utilization. They are complementary, as some kernels are memory-bound while others are compute-bound.

\subsection{Compared Methods}

We compare five methods spanning key design axes: general-purpose versus specialized models, iterative refinement and domain-specific training.

\textbf{AutoTriton}~\cite{li2025autotriton} is trained for Triton programming through supervised fine-tuning and reinforcement learning. We evaluate it in single-pass generation with its native prompting.

\textbf{GEAK}~\cite{wang2025geak} is an agentic framework with generator, evaluator, reflector, and optimizer modules. We use DeepSeek-V3.2-Chat as the base model, run three iterations at temperature 1.0, generate four candidates per round, and retain five best implementations as context in each round.

\textbf{KernelAgent}~\cite{kernelagent2025} is a multi-agent system based on a generate–verify–refine workflow, also using DeepSeek-V3.2-Chat. We bypass its Fuser pipeline (not applicable to single-operator tasks) and use its core generation API with 3 parallel workers, up to 5 refinement rounds each, at temperature 0.4.

\textbf{Claude} is a strong general-purpose model evaluated in single-pass generation mode.

\textbf{DeepSeek-Coder}~\cite{liu2024deepseek} is a general-purpose code model serving as a zero-specialization baseline. It is evaluated in single-pass generation mode.
\section{Methodology}  
\label{Sec:Methodology}

\subsection{System Pipeline}

\begin{figure*}[t]
    \centering
    \includegraphics[width=\textwidth]{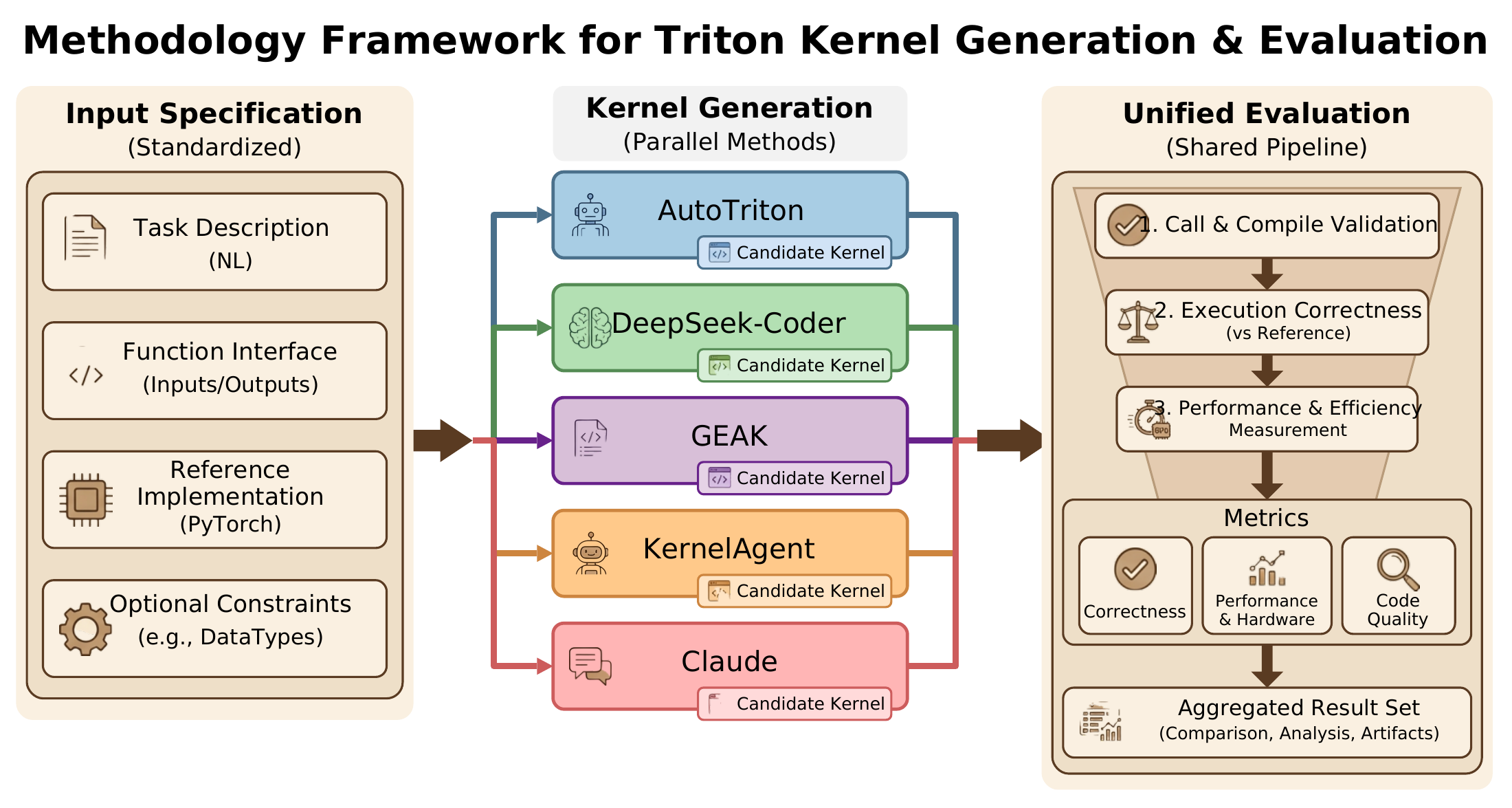}
    \caption{Multiple methods generate Triton kernels from a shared task specification, and are evaluated under a unified pipeline measuring correctness, efficiency, and code quality.}
    \label{fig:exp_pipeline}
\end{figure*}

Figure~\ref{fig:exp_pipeline} illustrates the full evaluation pipeline. For each task, \oursys provides a unified description including function interface, reference implementation, and task-specific constraints. Each method generates a candidate kernel through its dedicated adapter. All kernels are then evaluated by the same pipeline, with intermediate logs retained for error analysis.

\subsection{Benchmark Design}

\subsubsection{Task Organization}

\oursys contains 176 tasks across 15 categories: Activation, Convolution, Fusion, Index, LinearAlgebra, Loss, Math, MatrixMultiply, Normalization, Optimizer, Pooling, Quantization, Random, Reduce, and SpatialOps. Categories were assigned based on computational structure rather than operator type, enabling comparison of tasks with similar parallel execution requirements.

The benchmark includes two main coverage extensions beyond TritonBench-T. First, selected fp16, bf16, and int8 multi-precision variants test whether methods can generate kernels under low-precision constraints. Second, six quantization tasks in W8A8 and W4A16 settings test whether methods can implement manual quantization logic—including scale computation, explicit casting, and dequantization—without relying on high-level APIs.

\subsubsection{Correctness Protocol}

\oursys uses a two-stage correctness protocol designed to reject implementations that pass simple output comparisons by chance.

\textbf{Call Accuracy} checks whether generated code can be imported, compiled, and called correctly, and whether it satisfies task-level constraints. For quantization tasks, a static checker additionally rejects forbidden high-level quantization APIs and verifies the presence of manual quantization logic including scale computation and explicit casting operations.

\textbf{Execution Accuracy} checks whether outputs match the reference implementation across multiple input distributions. Each task is evaluated under two modes: a standard mode sampling inputs from $\mathcal{N}(0,1)$, and an outlier mode injecting amplified outliers (probability 0.1\%, scale factor 50) to expose implementations that pass on typical inputs but fail under distributional shift. For standard tasks, dtype-aware numerical tolerances are applied. For quantization tasks, three metrics must simultaneously satisfy task-specific thresholds: cosine similarity ($\geq 0.90$--$0.95$), L1 relative error ($\leq 0.05$--$0.10$), and RMSE ($\leq 0.10$--$0.15$).

\subsubsection{Performance and Code Quality Protocol}

Runtime is measured with triton.testing.do\_bench (25 warmup, 100 measurement runs, median reported). Speedup is computed against the PyTorch eager baseline.
Hardware efficiency is measured as $\max(\mathrm{IOU}, \mathrm{MFU})$ to evaluate each kernel against its dominant bottleneck. Code quality is assessed through Maintainability Index (MI) and Cyclomatic Complexity (CC).

Note that FLOP- and byte-based quantities are derived from a fixed task-level target model (the intended computation under an idealized implementation), and should be interpreted as normalized efficiency proxies rather than measurements of the actual executed instructions.
\section{Experiments} 
\label{Sec:Experiments}

\subsection{Experimental Setup}

We evaluate all five methods on six NVIDIA GPUs: RTX 5090, RTX 4090, A100-PCIE-40GB, H20, H800 PCIe, and L20, under a unified software stack (Python 3.11, CUDA 11.8, PyTorch 2.10.0+cu128, Triton 3.6.0). 
All cross-machine speedups are computed against PyTorch baselines remeasured on each target machine. All performance statistics are reported over semantically correct kernels only.

\subsection{Overall Results}
\label{sec:overall_results}

Table~\ref{tab:overall-main} reveals a sharp separation across success stages: compile success, semantic correctness, and useful acceleration are distinct outcomes that do not imply one another.

Two patterns stand out.
First, a large fraction of compiled kernels remain incorrect: although 64.2\% of KernelAgent-generated kernels compile successfully, only 10.8\% are correct, yielding a Correct/Compile conversion of 16.8\%.
Second, even the strongest methods remain far from reliable. GEAK yields the highest overall correctness at only 30.7\%, while Claude yields 22.7\%.
These results motivate shifting the analysis from aggregate ranking toward understanding where and why the success frontier breaks down.

\begin{table}[t]
\centering
\small
\setlength{\tabcolsep}{7pt}
\caption{Overall results on 176 tasks. Speedup and score are averaged over correct kernels only. Correct/Compile highlights how often compiled candidates actually preserve semantics.}
\begin{tabular}{lccccc}
\toprule
Method & Compile (\%) & Correct (\%) & Correct/Compile (\%) & Speedup & Score (\%) \\
\midrule
AutoTriton  & 36.4 & 17.0 & 46.9 & 1.35 & 60.7 \\
GEAK        & 68.8 & 30.7 & 44.6 & 1.15 & 50.0 \\
KernelAgent & 64.2 & 10.8 & 16.8 & 1.41 & 68.1 \\
Claude      & 45.5 & 22.7 & 50.0 & 1.26 & 54.2 \\
KernelLLM   & 1.7  & 0.0  & 0.0  & --   & 0.0 \\
\bottomrule
\end{tabular}
\label{tab:overall-main}
\end{table}

\subsection{Correctness Is Category-Structured}
\label{sec:category_correctness}

\begin{figure*}[t]
    \centering
    \includegraphics[width=0.88\textwidth]{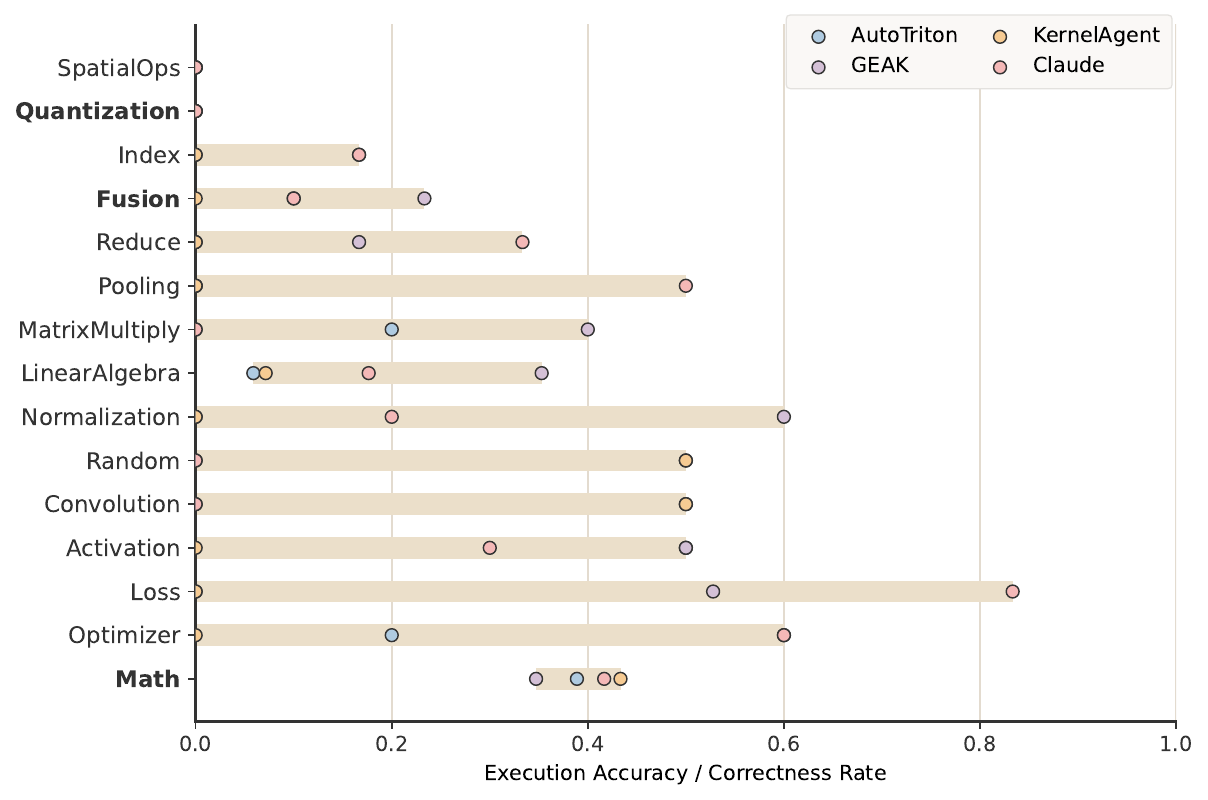}
    \caption{Category-Wise Correctness Across Methods.}
    \label{fig:semantic_correctness_category_boundary}
\end{figure*}

Figure~\ref{fig:semantic_correctness_category_boundary} reveals a striking pattern: correctness rates vary dramatically across categories---from near-zero in \texttt{SpatialOps} and \texttt{Quantization} to over 0.8 in \texttt{Loss}---while different methods tend to cluster within the same category band rather than separating across it. This suggests that task structure, not method identity, is the primary driver of correctness.

To quantify this, we fit task-level logistic attribution models using Pearson correlation of binary outcomes against method-indicator and category-indicator variables, reporting explained deviance as a measure of predictive power (excluding the near-zero DeepSeek-Coder baseline). Method identity and category identity explain nearly identical variance in compile success (5.18\% vs.\ 5.24\%), but for semantic correctness, category explains nearly three times more variance than method identity (9.4\% vs.\ 3.3\%). Correspondingly, adding a category on top of the method reduces deviance by 34.6, whereas adding the method on top of the category reduces it by only 13.0. While method design still matters at the executable stage, semantic correctness is primarily bounded by task structure.

Table~\ref{tab:category-conversion} further exposes where failures concentrate. Easy categories such as \texttt{Activation} and \texttt{Math} convert compiled candidates to correct kernels at rates of 46--56\%, while hard categories such as \texttt{Fusion} and \texttt{MatrixMultiply} remain near 25\%, and \texttt{Quantization} and \texttt{SpatialOps} reach 0\%.

Critically, these failures in hard categories are not caused by front-end syntax problems---their compile rates are non-trivial. To probe whether failure is instead reducible to code complexity, we compute several task-level static structure proxies and estimate their Pearson correlation with pooled correctness failure. Intermediate assignment count and fusion call count yield the strongest signal ($r \approx 0.21$), while cyclomatic complexity and a logical-span proxy yield only $r \approx 0.15$. Notably, all static proxies are more predictive of compile failure than of semantic failure. This confirms that hard-category failures represent a distinct semantic boundary, not merely harder syntax generation.

\begin{table}[t]
\centering
\caption{Representative category-level conversion from compilable code to semantically correct kernels, averaged over AutoTriton, GEAK, KernelAgent, and Claude.}
\small
\setlength{\tabcolsep}{8pt}
\begin{tabular}{lccc}
\toprule
Category & Avg. Compile (\%) & Avg. Correct (\%) & Correct/Compile (\%) \\
\midrule
Activation      & 70.0 & 32.5 & 46.4 \\
Math            & 72.2 & 40.3 & 55.8 \\
Fusion          & 43.8 & 10.8 & 24.8 \\
MatrixMultiply  & 60.0 & 15.0 & 25.0 \\
Quantization    & 41.7 & 0.0  & 0.0 \\
SpatialOps      & 25.0 & 0.0  & 0.0 \\
\bottomrule
\end{tabular}
\label{tab:category-conversion}
\end{table}

\subsection{Iterative Refinement Repairs Rather Than Optimizes}
\label{sec:iterative_refinement}

\begin{figure*}[t]
    \centering
    \includegraphics[width=0.90\textwidth]{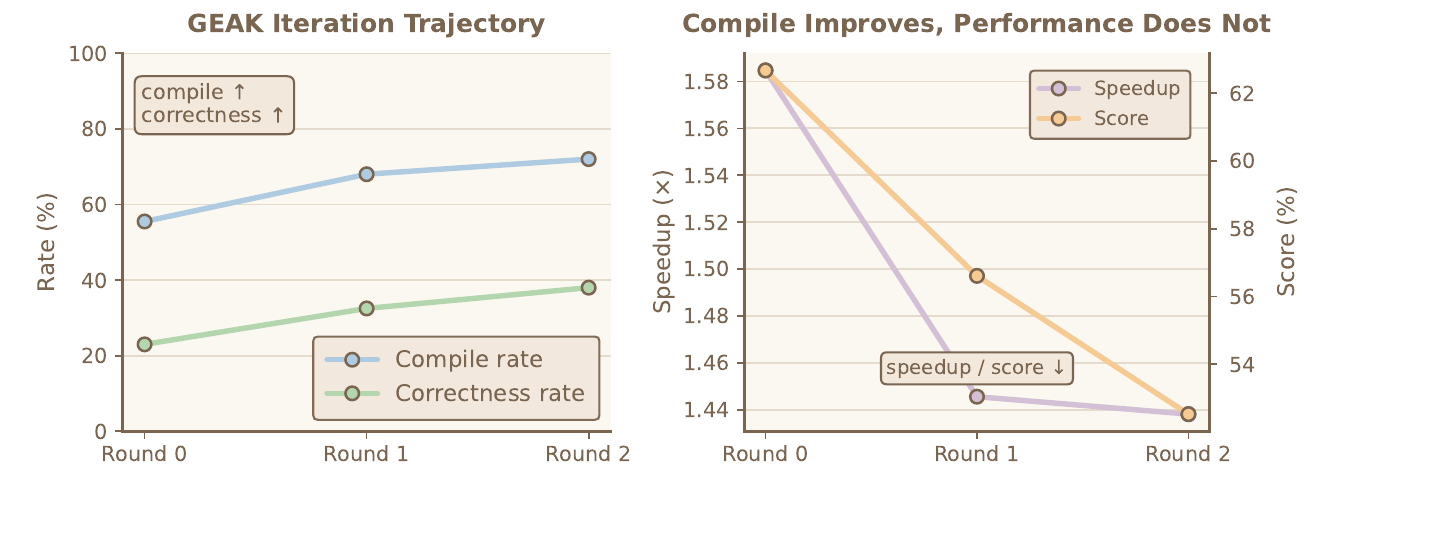}
    \caption{GEAK Iteration Trajectory.}
    \label{fig:geak_iteration_trajectory}
\end{figure*}

Figure~\ref{fig:geak_iteration_trajectory} illustrates a consistent
pattern: compile success rises from 52.3\% to 68.8\% and correctness from 18.2\%
to 30.7\%, but average speedup falls from 1.58$\times$ to 1.44$\times$ and score
from 62.7\% to 53.3\%. This performance decline will not be reversed by further iteration. KernelAgent shows the same pattern more starkly: many candidates compile, but few preserve semantics.

The performance drop is explained by the quality of newly rescued kernels. In GEAK 
rounds 0$\to$1, newly rescued correct kernels average only $1.16\times$ speedup 
(score 43.7\%), versus $1.58\times$ (score 62.9\%) for already-correct kernels. In 
rounds 1$\to$2, the gap persists: $1.32\times$ (score 35.5\%) versus $1.46\times$ 
(score 56.0\%). Analysis of 352 adjacent GEAK diffs confirms why: dominant edit 
types are no substantial change (102), mask fixes (101), delegated-op 
introduction/removal (65), and dtype/casting fixes (36), while optimization-oriented 
rewrites are rare. We analyze the structural reason for this repair bias in 
Insight~2 (\S\ref{sec:insight2}).

\subsection{High-Performance Kernel Generation Remains Challenging}

\begin{figure*}[t]
    \centering
    \includegraphics[width=0.97\textwidth]{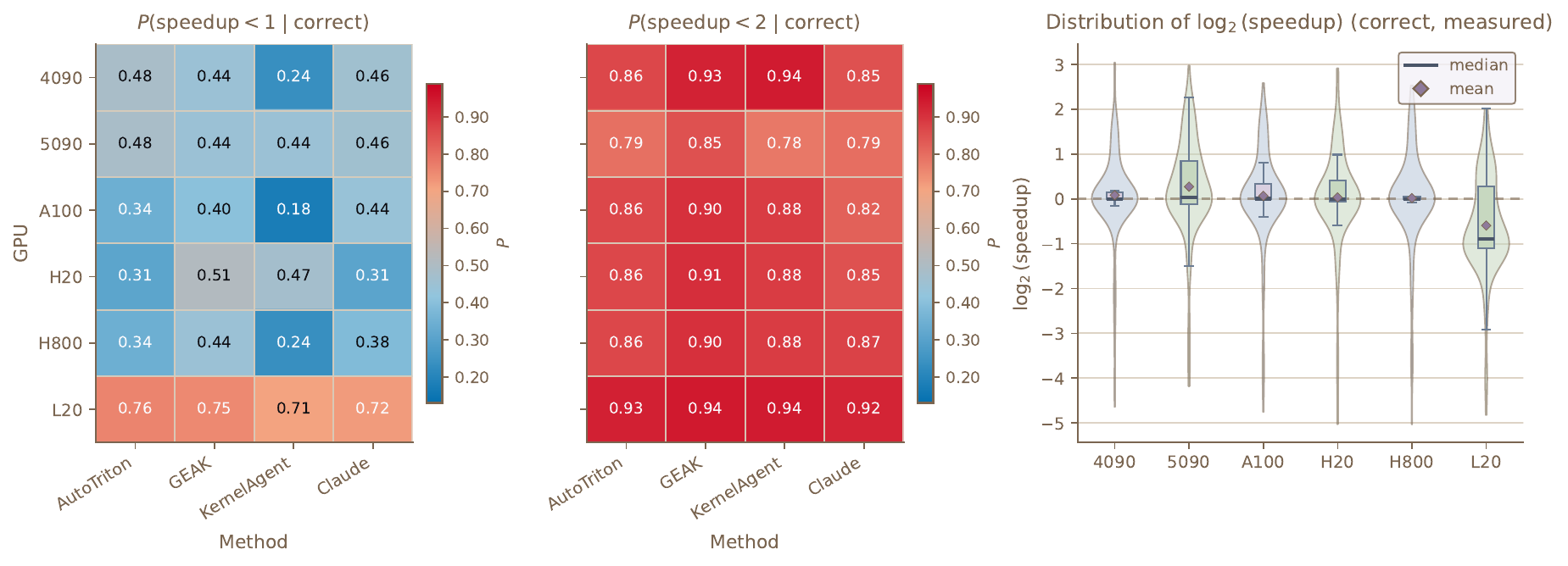}
    \caption{Cross-hardware speedup distribution and portability of correct kernels.}
    \label{fig:correct_speedup_machine_portability}
\end{figure*}

Semantic correctness is necessary, but alone insufficient for practical deployment.
Across all correct kernels, 46.6\% remain slower than eager PyTorch, and the pooled median speedup is only 1.0008$\times$. 
Cross-machine portability is also weak: the max/min speedup ratio has a median of $2.15\times$, a mean of $2.73\times$, and reaches $21.4\times$ in the worst case. Figure~\ref{fig:correct_speedup_machine_portability} shows that the fraction of correct kernels slower than PyTorch ranges from 18\% on A100 to 76\% on L20.

\subsection{Case Studies}

The following cases provide mechanistic illustrations of the three findings reported 
in Section~\ref{Sec:Insights}, grounding each claim in concrete kernel-level evidence.

\subsubsection{Case 1: Local, Single-Path Semantics Define the Success Ceiling}
\label{sec:case_success}

For the \texttt{logit} task, GEAK, Claude and AutoTriton all produce correct kernels achieving approximately $3.35\times$ speedup on RTX~4090. The task is structurally minimal: each output element depends on exactly one input element, clipping is purely local, and the kernel requires neither cross-block reduction nor inter-instance coordination.
Listing~\ref{lst:logit-autotriton-4090} shows a representative implementation. GEAK and Claude produce nearly identical kernels, differing only in whether clipping uses nested \texttt{tl.where} or \texttt{tl.minimum}/\texttt{tl.maximum}.

\begin{lstlisting}[style=papercode,label={lst:logit-autotriton-4090}]
    pid = tl.program_id(0)
    offsets = pid * XBLOCK + tl.arange(0, XBLOCK)
    mask = offsets < numel
    x = tl.load(input_ptr + offsets, mask=mask)
    if eps != 0.0:
        x = tl.where(x < eps, eps, x)
        x = tl.where(x > 1.0 - eps, 1.0 - eps, x)
    logit_x = tl.math.log(x / (1.0 - x))
    tl.store(output_ptr + offsets, logit_x, mask=mask)
\end{lstlisting}

This case establishes an upper bound: when data dependence is lane-local and the implementation path is near-template, most methods are reliable.

\subsubsection{Case 2: Non-Local Semantic Composition Fails at Correctness}
\label{sec:case_global_fail}

For \texttt{fused\_exp\_mean}, GEAK generates a kernel that compiles successfully, yet produces numerically incorrect outputs. The failure is not a syntax error, but a subtle interaction among padding semantics, a nonlinear transform, and a global reduction.

\begin{lstlisting}[style=papercode]
    x = tl.load(input_ptr + offset, mask=mask, other=0.0)
    exp_x = tl.math.exp(x.to(tl.float32))
    block_sum = tl.sum(exp_x, axis=0)
    if tl.program_id(0) == 0:
        tl.store(output_ptr, 0.0)
    tl.atomic_add(output_ptr, block_sum)
\end{lstlisting}

Each local idiom is individually correct. The error arises from their composition: masked-off lanes are padded with zero \emph{before} exponentiation, causing each to contribute \texttt{exp(0)\,=\,1} rather than 0 to the global reduction. The kernel violates the contract that only valid elements should participate in the mean---a failure invisible to local testing and only detectable at the full-tensor level.

\subsubsection{Case 3: Iterative Refinement Converges to Correct but Slow Kernels}
\label{sec:case_iter}

The GEAK trajectory on \texttt{Index/expand\_where} illustrates a repair that stops at correctness. The task requires \texttt{torch.where} over three operands that must be aligned under broadcasting before the predicate is applied. Round~0 fails to compile: the generator emits indexing logic the compiler rejects. Round~1 compiles but fails correctness: execution proceeds, yet the mapping from output positions to operand elements is wrong for mixed shapes. Round~2 fixes the mapping and passes the correctness gate---but records a speedup of only $0.076{\times}$.

\begin{lstlisting}[style=papercode]
    remaining = offsets
    for dim in range(output_ndim):
        multiplier = tl.load(output_multipliers_ptr + dim)
        coord = remaining // multiplier
        remaining = remaining % multiplier
        ...
\end{lstlisting}

Iterative feedback rewards correctness but provides insufficient signal about efficiency. As a result, the surviving kernel converges to an expensive implementation that recovers broadcast coordinates via radix decomposition and performs per-axis shape and stride lookups for each operand separately.
\section{Insights} \label{Sec:Insights}

\paragraph{Insight 1: Global-Contract Semantic Failure.}
As shown in Section~\ref{sec:category_correctness}, task category explains nearly three times more variance in semantic correctness than method identity (9.4\% vs.\ 3.3\%), and the correctness gap between easy and hard categories persists despite non-trivial compile rates---ruling out front-end syntax as the primary cause. Static complexity proxies correlate only weakly with failure ($r \leq 0.21$), confirming that the boundary is not reducible to code length or branching depth.
Instead, failures concentrate on tasks where correctness depends on maintaining consistent tensor semantics across different dimensions, memory layouts, and parallel program instances. These constraints span the entire kernel and are therefore difficult to recover through local edits alone.
Case~\ref{sec:case_global_fail} concretizes this mechanism: models produce individually correct Triton idioms while violating the global contract those idioms must collectively satisfy. 
Case~\ref{sec:case_success} provides the contrasting bound: when data dependence is lane-local, no such global contract exists, and all methods are reliably correct.

\paragraph{Insight 2: Repair-Biased Iterative Refinement.}
\label{sec:insight2}

As shown in Section~\ref{sec:iterative_refinement}, iterative refinement reliably expands compilability and correctness, but kernel performance often fails to improve and can even degrade across iterations.
The edit distribution over 352 GEAK diffs confirms that current loops operate primarily as repair mechanisms, with performance-oriented rewrites being rare.
The underlying asymmetry is structural: repair responds to explicit, local error signals (compilation errors, shape mismatches, failing outputs), while performance improvement requires plan-level decisions about tiling, memory layout, and kernel boundaries, which are not recoverable from the feedback available in current iterative pipelines. Case~\ref{sec:case_iter} illustrates this behavior in a concrete setting. As a result, iterative refinement reliably converges to semantically correct implementations, but not to efficient ones. The feedback signal that drove convergence provided insufficient information about this cost, and the iterative process offers no effective means to address it.

\paragraph{Insight 3: Performance as an Unsolved Frontier.}
Among semantically correct kernels, 46.6\% remain slower than eager PyTorch, and the pooled median speedup is only $1.0008\times$ (Section~\ref{sec:overall_results}). 
Cross-machine compounds this: the max/min speedup ratio reaches $21.4\times$ in the worst case, indicating that correct kernels are often hardware-specific rather than generally efficient.
Taken together with Insights~1 and~2, this suggests that correctness and performance represent distinct frontiers: current methods have partially crossed the correctness boundary, but closing the performance gap will likely require qualitatively different mechanisms, such as explicit hardware-aware search or performance-signal feedback, rather than refinements to existing correctness-driven pipelines.

\section{Conclusion}

We introduce \oursys to characterize the capability boundary of LLM-based Triton kernel generation through category-aware evaluation across 176 tasks on six GPUs. Three insights emerge.
The correctness boundary is category-structured and driven by non-local semantic coordination.
Iterative refinement is repair-biased: it expands the feasible set but introduces weaker candidates, the edit distribution is dominated by local fixes.
Performance validity remains a distinct and unsolved challenge.

Together, these results indicate that the capability boundary of current LLM-based kernel generation is not a single wall but a sequence of distinct barriers - compilability, semantic correctness, hardware efficiency and performance portability - each requiring different mechanisms to clear.
Prompt engineering and iterative refinement are well-suited to compilability but structurally insufficient for the rest.
Progress will likely require mechanisms for reasoning about global tensor contracts and parallel reduction semantics, training signals that reward numerical fidelity rather than surface resemblance, and efficiency-aware generation with explicit hardware cost feedback.
We release \oursys together with the iteration-level transition pairs collected during evaluation to support research in these directions.

\newpage
\bibliography{main}

@article{zhang2025sageattention3,
  title={Sageattention3: Microscaling fp4 attention for inference and an exploration of 8-bit training},
  author={Zhang, Jintao and Wei, Jia and Zhang, Pengle and Xu, Xiaoming and Huang, Haofeng and Wang, Haoxu and Jiang, Kai and Zhu, Jun and Chen, Jianfei},
  journal={arXiv preprint arXiv:2505.11594},
  year={2025}
}

@article{zhang2025sageattention2++,
  title={SageAttention2++: A More Efficient Implementation of SageAttention2},
  author={Zhang, Jintao and Xu, Xiaoming and Wei, Jia and Huang, Haofeng and Zhang, Pengle and Xiang, Chendong and Zhu, Jun and Chen, Jianfei},
  journal={arXiv preprint arXiv:2505.21136},
  year={2025}
}

@article{zhang2025spargeattn,
  title={Spargeattn: Accurate sparse attention accelerating any model inference},
  author={Zhang, Jintao and Xiang, Chendong and Huang, Haofeng and Wei, Jia and Xi, Haocheng and Zhu, Jun and Chen, Jianfei},
  journal={arXiv preprint arXiv:2502.18137},
  year={2025}
}

@article{zhang2024sageattention2,
  title={Sageattention2: Efficient attention with thorough outlier smoothing and per-thread int4 quantization},
  author={Zhang, Jintao and Huang, Haofeng and Zhang, Pengle and Wei, Jia and Zhu, Jun and Chen, Jianfei},
  journal={arXiv preprint arXiv:2411.10958},
  year={2024}
}

@inproceedings{zhangsageattention,
  title={SageAttention: Accurate 8-Bit Attention for Plug-and-play Inference Acceleration},
  author={Zhang, Jintao and Zhang, Pengle and Zhu, Jun and Chen, Jianfei and others},
  booktitle={The Thirteenth International Conference on Learning Representations}
}

@inproceedings{
shah2024flashattention,
title={FlashAttention-3: Fast and Accurate Attention with Asynchrony and Low-precision},
author={Jay Shah and Ganesh Bikshandi and Ying Zhang and Vijay Thakkar and Pradeep Ramani and Tri Dao},
booktitle={The Thirty-eighth Annual Conference on Neural Information Processing Systems},
year={2024}
}

@inproceedings{tillet2019triton,
  title={Triton: an intermediate language and compiler for tiled neural network computations},
  author={Tillet, Philippe and Kung, Hsiang-Tsung and Cox, David},
  booktitle={Proceedings of the 3rd ACM SIGPLAN International Workshop on Machine Learning and Programming Languages},
  pages={10--19},
  year={2019}
}

@article{liu2024deepseek,
  title={Deepseek-v2: A strong, economical, and efficient mixture-of-experts language model},
  author={Liu, Aixin and Feng, Bei and Wang, Bin and Wang, Bingxuan and Liu, Bo and Zhao, Chenggang and Dengr, Chengqi and Ruan, Chong and Dai, Damai and Guo, Daya and others},
  journal={arXiv preprint arXiv:2405.04434},
  year={2024}
}

@article{zhang2025sla,
  title={SLA: Beyond Sparsity in Diffusion Transformers via Fine-Tunable Sparse-Linear Attention},
  author={Zhang, Jintao and Wang, Haoxu and Jiang, Kai and Yang, Shuo and Zheng, Kaiwen and Xi, Haocheng and Wang, Ziteng and Zhu, Hongzhou and Zhao, Min and Stoica, Ion and others},
  journal={arXiv preprint arXiv:2509.24006},
  year={2025}
}

@article{liao2025kernelevolve,
  title={Kernelevolve: Scaling agentic kernel coding for heterogeneous ai accelerators at meta},
  author={Liao, Gang and Qin, Hongsen and Wang, Ying and Golden, Alicia and Kuchnik, Michael and Yetim, Yavuz and Ang, Jia Jiunn and Fu, Chunli and He, Yihan and Hsia, Samuel and others},
  journal={arXiv preprint arXiv:2512.23236},
  year={2025}
}

@article{wang2025geak,
  title={Geak: Introducing triton kernel ai agent \& evaluation benchmarks},
  author={Wang, Jianghui and Joshi, Vinay and Majumder, Saptarshi and Chao, Xu and Ding, Bin and Liu, Ziqiong and Brahma, Pratik Prabhanjan and Li, Dong and Liu, Zicheng and Barsoum, Emad},
  journal={arXiv preprint arXiv:2507.23194},
  year={2025}
}

@article{woo2025tritonrl,
  title={Tritonrl: Training llms to think and code triton without cheating},
  author={Woo, Jiin and Zhu, Shaowei and Nie, Allen and Jia, Zhen and Wang, Yida and Park, Youngsuk},
  journal={arXiv preprint arXiv:2510.17891},
  year={2025}
}

@article{li2025autotriton,
  title={Autotriton: Automatic triton programming with reinforcement learning in llms},
  author={Li, Shangzhan and Wang, Zefan and He, Ye and Li, Yuxuan and Shi, Qi and Li, Jianling and Hu, Yonggang and Che, Wanxiang and Han, Xu and Liu, Zhiyuan and others},
  journal={arXiv preprint arXiv:2507.05687},
  year={2025}
}

@article{ouyang2025kernelbench,
  title={Kernelbench: Can llms write efficient gpu kernels?},
  author={Ouyang, Anne and Guo, Simon and Arora, Simran and Zhang, Alex L and Hu, William and R{\'e}, Christopher and Mirhoseini, Azalia},
  journal={arXiv preprint arXiv:2502.10517},
  year={2025}
}

@article{wen2025multikernelbench,
  title={Multikernelbench: A multi-platform benchmark for kernel generation},
  author={Wen, Zhongzhen and Zhang, Yinghui and Li, Zhong and Liu, Zhongxin and Xie, Linna and Zhang, Tian},
  journal={arXiv e-prints, pp. arXiv--2507},
  year={2025}
}

@inproceedings{li2025tritonbench,
  title={Tritonbench: Benchmarking large language model capabilities for generating triton operators},
  author={Li, Jianling and Li, Shangzhan and Gao, Zhenye and Shi, Qi and Li, Yuxuan and Wang, Zefan and Huang, Jiacheng and WangHaojie, WangHaojie and Wang, Jianrong and Han, Xu and others},
  booktitle={Findings of the Association for Computational Linguistics: ACL 2025},
  pages={23053--23066},
  year={2025}
}

@article{gong2025large,
  title={From large to small: Transferring cuda optimization expertise via reasoning graph},
  author={Gong, Junfeng and Wei, Zhiyi and Chen, Junying and Liu, Cheng and Li, Huawei},
  journal={arXiv preprint arXiv:2510.19873},
  year={2025}
}

@article{lange2025towards,
  title={Towards robust agentic cuda kernel benchmarking, verification, and optimization},
  author={Lange, Robert Tjarko and Sun, Qi and Prasad, Aaditya and Faldor, Maxence and Tang, Yujin and Ha, David},
  journal={arXiv preprint arXiv:2509.14279},
  year={2025}
}

@article{dong2025stark,
  title={Stark: Strategic team of agents for refining kernels},
  author={Dong, Juncheng and Yang, Yang and Liu, Tao and Wang, Yang and Qi, Feng and Tarokh, Vahid and Rangadurai, Kaushik and Yang, Shuang},
  journal={arXiv preprint arXiv:2510.16996},
  year={2025}
}

@misc{kernelagent2025,
  author={Wang, Laura and others},
  title={KernelFalcon: Deep Agent Architecture for Autonomous GPU Kernel Generation},
  howpublished={\url{https://pytorch.org/blog/kernelagent-hardware-guided-gpu-kernel-optimization-via-multi-agent-orchestration/}},
  note={PyTorch Blog, November 2025}
}

@article{yu2026towards,
  title={Towards Automated Kernel Generation in the Era of LLMs},
  author={Yu, Yang and Zang, Peiyu and Tsai, Chi Hsu and Wu, Haiming and Shen, Yixin and Zhang, Jialing and Wang, Haoyu and Xiao, Zhiyou and Shi, Jingze and Luo, Yuyu and others},
  journal={arXiv preprint arXiv:2601.15727},
  year={2026}
}

@article{chen2021evaluating,
  title={Evaluating large language models trained on code},
  author={Chen, Mark and Tworek, Jerry and Jun, Heewoo and Yuan, Qiming and Pinto, Henrique Ponde De Oliveira and Kaplan, Jared and Edwards, Harri and Burda, Yuri and Joseph, Nicholas and Brockman, Greg and others},
  journal={arXiv preprint arXiv:2107.03374},
  year={2021}
}

@article{williams2009roofline,
  title={Roofline: an insightful visual performance model for multicore architectures},
  author={Williams, Samuel and Waterman, Andrew and Patterson, David},
  journal={Communications of the ACM},
  volume={52},
  number={4},
  pages={65--76},
  year={2009},
  publisher={ACM New York, NY, USA}
}

@article{zhang2026sla2,
  title        = {{SLA2: Sparse-Linear Attention with Learnable Routing and QAT}},
  author={Zhang, Jintao and Wang, Haoxu and Jiang, Kai and Zheng, Kaiwen and Jiang, Youhe and Stoica, Ion and Chen, Jianfei and Zhu, Jun and Gonzalez, Joseph E.},
  year         = {2026}
}

@article{zhang2025turbodiffusion,
  title={TurboDiffusion: Accelerating Video Diffusion Models by 100-200 Times},
  author={Zhang, Jintao and Zheng, Kaiwen and Jiang, Kai and Wang, Haoxu and Stoica, Ion and Gonzalez, Joseph E and Chen, Jianfei and Zhu, Jun},
  journal={arXiv preprint arXiv:2512.16093},
  year={2025}
}

@article{zhangefficient,
  title={Efficient Attention Methods: Hardware-efficient, Sparse, Compact, and Linear Attention},
  author={Zhang, Jintao and Su, Rundong and Liu, Chunyu and Wei, Jia and Wang, Ziteng and Wang, Haoxu and Zhang, Pengle and Jiang, Huiqiang and Huang, Haofeng and Xiang, Chendong and others}
}

@article{zhang2026sagebwd,
  title={SageBwd: A Trainable Low-bit Attention},
  author={Zhang, Jintao and Chen, Marco and Wang, Haoxu and Jiang, Kai and Stoica, Ion and Gonzalez, Joseph E and Chen, Jianfei and Zhu, Jun},
  journal={arXiv preprint arXiv:2603.02170},
  year={2026}
}
\bibliographystyle{tmlr}

\newpage

\appendix
\appendix

\section*{Technical Appendices and Supplementary Material}

\noindent
- Section~\hyperref[app:benchmark]{A}: Benchmark and Evaluation Details \\
- Section~\hyperref[app:detail_results]{B}: Detailed Results \\
- Section~\hyperref[app:quantization-details]{C}: Quantization Details \\
- Section~\hyperref[app:analysis]{D}: Analysis: Why LLMs Cannot Reliably Generate High-Performance Kernels \\
- Section~\hyperref[app:artifacts]{E}: Artifacts: Transition Pairs for Repair and Optimization Analysis

\section{Benchmark and Evaluation Details}
\label{app:benchmark}

\subsection{Task List and Category Taxonomy}

\oursys contains 176 tasks spanning 15 fine-grained categories. Rather than organizing tasks by operator type, we group them by the \emph{type of knowledge required to produce a correct implementation}, enabling category-level analysis of systematic failure modes.

\paragraph{Direct specification tasks (Activation, Math).}
\textbf{Activation} (10) and \textbf{Math} (36) are primarily defined by explicit formulas or standard library semantics (e.g.\@, ReLU, softmax). Correctness largely reduces to faithful translation of the specification, with occasional subtleties in composition or numerical stability.

\paragraph{Parallel aggregation structures (Reduce, Pooling, Normalization).}
\textbf{Reduce} (6), \textbf{Pooling} (2), and \textbf{Normalization} (5) depend on aggregation over a defined scope (e.g., reduction axis or normalization domain). 
Errors typically arise from incorrect scope or inconsistent statistics; some variants extend to matrix-level aggregation (e.g., spectral normalization).

\paragraph{Structured multi-operand computation (MatrixMultiply, LinearAlgebra).}
\textbf{MatrixMultiply} (10) centers on structured two-operand computation, often combined with additional reductions. \textbf{LinearAlgebra} (17) involves decompositions (SVD, QR, LU) with coupled multi-output constraints (e.g., $A = U S V^H$), requiring consistency across outputs.

\paragraph{Indexed and spatial computation (Convolution, Index, SpatialOps).}
\textbf{Convolution} (2), \textbf{Index} (6), and \textbf{SpatialOps} (3) are dominated by address computation rather than value arithmetic. Convolution requires boundary-aware sliding windows; Index tasks involve gather/scatter or masked selection; SpatialOps rely on coordinate mapping and interpolation.

\paragraph{Compositional kernels (Fusion).}
\textbf{Fusion} (60) composes multiple operations within a single kernel. Correctness depends on preserving invariants across operation boundaries, especially when combining elementwise, reduction, and normalization steps.

\paragraph{Semantic contract categories (Loss, Optimizer, Random, Quantization).}
These categories require knowledge beyond direct formulas. \textbf{Loss} and \textbf{Optimizer} involve API-level contracts (e.g., reduction modes, state updates). \textbf{Random} includes both stochastic sampling and deterministic tensor factories (e.g., \texttt{logspace}), requiring consistency under seeding and device semantics. \textbf{Quantization} follows an approximation contract, evaluated by multiple precision metrics rather than exact equality.

\paragraph{Discussion.}
These categories reflect different types of knowledge: explicit specifications, structured parallel patterns, compositional reasoning, and contract-level semantics. 
We hypothesize that the latter are less consistently represented in pretraining data, which may explain their lower correctness in Section~\ref{sec:category_correctness}.

\begin{figure*}[t]
    \centering
    \includegraphics[width=0.95\textwidth]{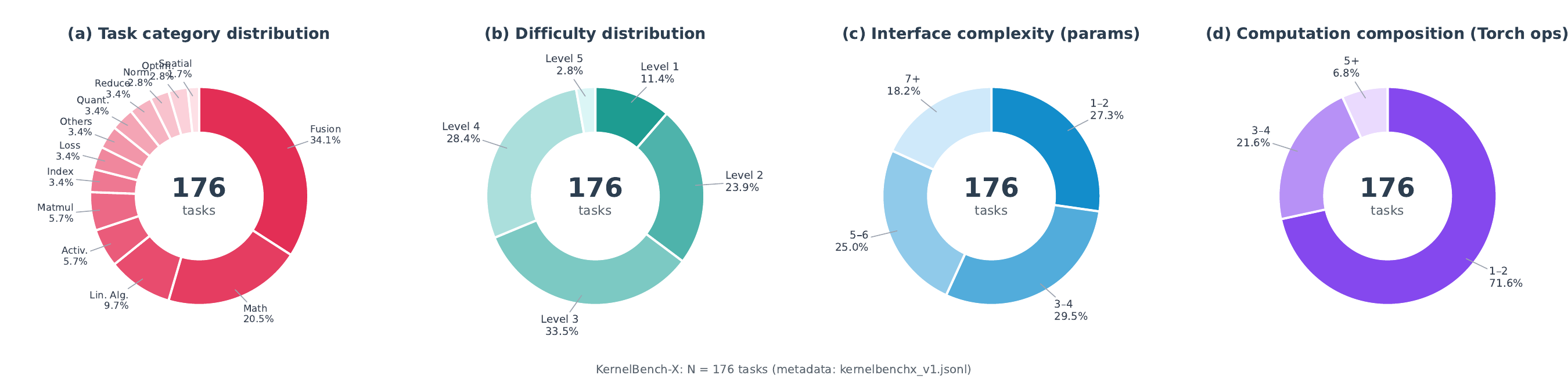}
    \caption{Overview of \oursys benchmark structure.}
    \label{fig:benchmark_overview}
\end{figure*}

\subsection{Call Accuracy}

A task is counted as passing the call stage only if the generated prediction is non-empty, exports a valid kernel entry according to the benchmark’s AST-based kernel-entry check, and executes successfully under the call harness. For Quantization tasks, an additional static quantization check is applied at this stage.

\subsection{Execution Accuracy}

Execution accuracy compares generated outputs against the benchmark reference implementation under a shared random seed. Both sides must expose a test\_results object, and comparison is recursive. Exact shape and dtype agreement are required before numerical comparison for all tasks.

\subsection{Performance Metrics and Aggregation}

Per-task speedup is the ratio of total reference runtime to total generated kernel runtime across all benchmarked inputs for a given task. Two aggregate speedup notions are available: a global ratio of total times, and an arithmetic mean of per-task speedups. These quantities are not interchangeable and may diverge substantially under outliers. The paper primarily reports the latter.

\section{Detailed Results}
\label{app:detail_results}

\subsection{Full Category-Level Results}

Tables~\ref{tab:appendix-full-category-correct} and \ref{tab:appendix-full-category-compile} report the full correctness and compile rates for all methods and categories.

\begin{table*}[t]
\centering
\small
\setlength{\tabcolsep}{5pt}
\caption{Per-category semantic correctness rates (\%) for all benchmark categories.}
\begin{tabular}{lcccccc}
\toprule
Category & \#Tasks & AutoTriton & GEAK & KernelAgent & Claude & KernelLLM \\
\midrule
Activation     & 10 & 40.0 & 50.0 & 0.0  & 40.0 & 0.0 \\
Convolution    & 2  & 0.0  & 0.0  & 50.0 & 0.0  & 0.0 \\
Fusion         & 60 & 10.0 & 23.3 & 0.0  & 10.0 & 0.0 \\
Index          & 6  & 16.7 & 16.7 & 0.0  & 16.7 & 0.0 \\
LinearAlgebra  & 17 & 5.9  & 35.3 & 5.9  & 17.6 & 0.0 \\
Loss           & 6  & 16.7 & 50.0 & 0.0  & 33.3 & 0.0 \\
Math           & 36 & 30.6 & 36.1 & 44.4 & 50.0 & 0.0 \\
MatrixMultiply & 10 & 20.0 & 40.0 & 0.0  & 0.0  & 0.0 \\
Normalization  & 5  & 40.0 & 60.0 & 0.0  & 20.0 & 0.0 \\
Optimizer      & 5  & 20.0 & 60.0 & 0.0  & 20.0 & 0.0 \\
Pooling        & 2  & 50.0 & 0.0  & 0.0  & 0.0  & 0.0 \\
Quantization   & 6  & 0.0  & 0.0  & 0.0  & 0.0  & 0.0 \\
Random         & 2  & 0.0  & 50.0 & 50.0 & 50.0 & 0.0 \\
Reduce         & 6  & 0.0  & 16.7 & 0.0  & 33.3 & 0.0 \\
SpatialOps     & 3  & 0.0  & 0.0  & 0.0  & 0.0  & 0.0 \\
\bottomrule
\end{tabular}
\label{tab:appendix-full-category-correct}
\end{table*}

\begin{table*}[t]
\centering
\small
\setlength{\tabcolsep}{5pt}
\caption{Per-category compile success rates (\%) for all benchmark categories.}
\begin{tabular}{lcccccc}
\toprule
Category & \#Tasks & AutoTriton & GEAK & KernelAgent & Claude & KernelLLM \\
\midrule
Activation     & 10 & 80.0  & 60.0  & 60.0  & 50.0  & 0.0 \\
Convolution    & 2  & 50.0  & 0.0   & 50.0  & 0.0   & 0.0 \\
Fusion         & 60 & 30.0  & 68.3  & 40.0  & 36.7  & 1.7 \\
Index          & 6  & 33.3  & 50.0  & 16.7  & 66.7  & 0.0 \\
LinearAlgebra  & 17 & 11.8  & 64.7  & 29.4  & 41.2  & 0.0 \\
Loss           & 6  & 16.7  & 83.3  & 16.7  & 83.3  & 0.0 \\
Math           & 36 & 58.3  & 75.0  & 63.9  & 61.1  & 0.0 \\
MatrixMultiply & 10 & 40.0  & 90.0  & 50.0  & 30.0  & 0.0 \\
Normalization  & 5  & 80.0  & 60.0  & 40.0  & 40.0  & 0.0 \\
Optimizer      & 5  & 20.0  & 80.0  & 40.0  & 60.0  & 0.0 \\
Pooling        & 2  & 100.0 & 50.0  & 0.0   & 50.0  & 0.0 \\
Quantization   & 6  & 0.0   & 50.0  & 83.3  & 33.3  & 0.0 \\
Random         & 2  & 0.0   & 100.0 & 50.0  & 50.0  & 0.0 \\
Reduce         & 6  & 0.0   & 66.7  & 16.7  & 50.0  & 0.0 \\
SpatialOps     & 3  & 0.0   & 66.7  & 0.0   & 0.0   & 0.0 \\
\bottomrule
\end{tabular}
\label{tab:appendix-full-category-compile}
\end{table*}

\subsection{Static Structure Proxies}

To probe whether the correctness boundary is reducible to code complexity, we computed task-level static structure proxies and their Pearson correlations with pooled correctness failure. Results are shown in Table~\ref{tab:appendix-structure-proxies}.

All proxies correlate only modestly with correctness failure, and are more predictive of compile failure than semantic failure. This confirms that the benchmark’s correctness boundary is structural but not reducible to superficial measures of reference-code complexity.

\begin{table}[h]
\centering
\small
\caption{Static structure proxies versus pooled correctness failure.}
\begin{tabular}{lc}
\toprule
Metric & Pearson \(r\) with pooled correctness failure \\
\midrule
Cyclomatic complexity (max) & 0.1454 \\
Logical span proxy          & 0.1453 \\
Intermediate assignment count & 0.2114 \\
Fusion torch-call count       & 0.2114 \\
\bottomrule
\end{tabular}
\label{tab:appendix-structure-proxies}
\end{table}

\section{Quantization Details}
\label{app:quantization-details}

\oursys contains six quantization tasks: matmul\_w8a8, bmm\_w8a8, conv2d\_w8a8, layernorm\_w8a8, attention\_w8a8, and linear\_w4a16. Generated code is rejected if it relies on forbidden high-level quantization APIs; instead, kernels must implement explicit quantization logic such as scale computation and discretization. Unlike standard benchmark tasks, all six quantization tasks use custom execution-stage precision thresholds (Table~\ref{tab:appendix-quant-thresholds}).

\begin{table}[h]
\centering
\small
\caption{Custom precision thresholds for the six active quantization tasks.}
\begin{tabular}{lcccc}
\toprule
Task & Scheme & Cosine \(\ge\) & \(L_1\) Relative \(\le\) & RMSE \(\le\) \\
\midrule
matmul\_w8a8    & W8A8  & 0.95 & 0.05 & 0.10 \\
bmm\_w8a8       & W8A8  & 0.95 & 0.05 & 0.10 \\
conv2d\_w8a8    & W8A8  & 0.95 & 0.05 & 0.10 \\
layernorm\_w8a8 & W8A8  & 0.95 & 0.05 & 0.10 \\
attention\_w8a8 & W8A8  & 0.90 & 0.10 & 0.15 \\
linear\_w4a16   & W4A16 & 0.90 & 0.10 & 0.15 \\
\bottomrule
\end{tabular}
\label{tab:appendix-quant-thresholds}
\end{table}

Quantization should not be interpreted as merely another difficult category. Its 0\% correctness despite non-trivial compilation reveals a distinct semantic boundary: models do not reliably treat the numerical contract itself as part of the computation to be preserved.

\section{Analysis: Why LLMs Cannot Reliably Generate High-Performance Kernels}
\label{app:analysis}

This appendix analyzes the systemic reasons why current LLM-based methods fail to produce hardware-efficient Triton kernels.
We organize the analysis around three structural factors: training data, prompt construction and iterative feedback.

\subsection{Training Data Lacks Performance Grounding}

Base LLMs are trained on code corpora in which performance is not annotated---source code is treated as semantic text, not as a description of hardware behavior.
A model trained in this way can learn to produce code that \emph{looks like} an efficient kernel without acquiring any representation of why it is fast or slow, or under what hardware conditions that changes. This is consistent with findings that LLMs trained on code develop strong syntactic priors but weak hardware-behavioral representations ~\cite{chen2021evaluating}.

Methods such as AutoTriton try to address this limitation by introducing execution-based supervision during training. However, this supervision remains correctness-oriented rather than performance-oriented: generated kernels are rewarded for satisfying execution and test constraints, not for achieving robust efficiency across hardware platforms. As a result, models can learn to imitate efficient-looking implementations without acquiring a transferable understanding of hardware-dependent performance behavior.
This limitation is directly reflected in our results. Among semantically correct kernels, cross-hardware speedup variance reaches $21.4\times$ in the worst case, while the fraction of correct kernels that are slower than PyTorch ranges from 18\% on A100 to 76\% on L20 (Section~\ref{sec:overall_results}). AutoTriton itself achieves a mean speedup of only $1.35\times$ and exhibits cross-platform variance comparable to methods without execution-based supervision, suggesting that correctness-oriented training signals do not translate into robust hardware-aware optimization on unseen platforms.

\subsection{Prompt Construction Omits Hardware Context}

None of the evaluated methods are designed to take hardware information as an explicit input. This is a structural limitation rather than an implementation oversight.
Without hardware context, a model cannot reason about whether a given tiling strategy will fit in shared memory, whether a particular num\_warps setting will cause register spilling, or whether a memory access pattern will achieve coalescing on the target device.
Generated kernels therefore implicitly target average or prototypical hardware, and their performance degrades unpredictably across platforms.

\subsection{Iterative Feedback Cannot Drive Performance Optimization}

Current iterative pipelines provide two kinds of feedback: compilation errors and correctness failures. Both are explicit and local, which is why iterative refinement reliably expands compilability and correctness. But performance optimization requires knowing whether the kernel is memory-bound or compute-bound and how the hardware scheduler is allocating execution resources~\cite{williams2009roofline}---none of which is recoverable from a correctness failure or compiler error.

Our edit-level analysis of adjacent GEAK diffs confirms this asymmetry directly (Section~\ref{sec:iterative_refinement}): as refinement rescues additional correct kernels, average speedup falls, because newly rescued kernels consistently underperform those that were correct from the start. Case~3 is one such example. Substantive performance gains require non-local structural changes---retiling, restructuring reductions, reconsidering kernel boundaries---that are unreachable by local neighborhood search from a correct but inefficient starting point. LLM-based refinement, operating over token-level patterns without access to hardware-level cost signals, cannot drive this kind of search.

\subsection{Potential Improvement Directions}

Two directions may help address these limitations.
First, profile-guided hyperparameter search can tune hardware-sensitive parameters---block size, warp count, pipeline stages---on top of an already correct kernel, finding a strong configuration without exhaustive search. 
Second, hardware-aware training could expose models to hardware specifications together with cross-platform performance outcomes, teaching them how implementation choices interact with different processors. Such a model could then interpret and act on profiling signals---roofline analysis, bandwidth utilization, compute utilization---during iterative refinement. A model without this training cannot meaningfully use such signals even when they are provided.

\section{Artifacts: Transition Pairs for Repair and Optimization Analysis}
\label{app:artifacts}

Beyond aggregate results, we collect an iteration-level artifact 
set centered on \emph{transition events} between adjacent GEAK rounds.

\paragraph{Correctness transitions.}
The primary artifact family records cases where results change from incorrect to correct across successive rounds.
Each record includes before/after pass indicators, runtime and speedup when available, aligned error text, and a unified diff. This provides an explicit error $\rightarrow$ patch $\rightarrow$ pass chain suitable for local repair modeling.

\textbf{Performance events} record rounds with execution time improvement over the previous round. \textbf{Regression events} record rounds where a previously passing metric fails. Both are relatively infrequent in our experiment.

\end{document}